\let\origvec\vec
\documentclass[runningheads]{llncs}

\let\vec\origvec
\usepackage{hyperref}
\usepackage{graphicx}
\usepackage{amsmath}
\usepackage{bm}
\usepackage{tabularx}
\usepackage{siunitx}
\usepackage{color, colortbl}
\usepackage{multirow}
\usepackage[utf8]{inputenc}
\usepackage{floatrow}

\usepackage{xcolor}

\definecolor{DarkGray}{gray}{0.85}
\definecolor{Gray}{gray}{0.90}
\usepackage{tikz,pgfplots}
\usetikzlibrary{plotmarks,shapes,arrows}
\pgfplotsset{compat=newest} 
\pgfkeys{/pgf/number format/.cd,1000 sep={}}

\definecolor{LightCyan}{rgb}{0.88,1,1}
\definecolor{LightRed}{rgb}{1,0.7,0.4}
\definecolor{LightGray}{rgb}{0.86,0.86,0.86}
\definecolor{LightBlue}{rgb}{0.6,0.8,1}
\definecolor{LightYellow}{rgb}{0.98,0.89,0.46}
\begin{document}
\title{CubiCasa5K: A Dataset and an Improved Multi-Task Model for Floorplan Image Analysis}
\titlerunning{CubiCasa5K: A Dataset and Framework for Floorplan Image Analysis}
%
\author{Ahti Kalervo\inst{1} \and
Juha Ylioinas\inst{1} \and
Markus H\"aiki\"o\inst{2} \and
Antti Karhu\inst{2} \and
Juho Kannala\inst{1}}

\authorrunning{A. Kalervo et al.}
%
\institute{Department of Computer Science, Aalto University, Espoo, Finland \\
\email{\{firstname.lastname\}@aalto.fi}
\and CubiCasa Inc., Oulu, Finland \\
\email{\{firstname.lastname\}@cubicasa.com}}
\maketitle              
\begin{abstract}
Better understanding and modelling of building interiors and the emergence of more impressive AR/VR technology has brought up the need for automatic parsing of floorplan images. However, there is a clear lack of representative datasets to investigate the problem further. To address this shortcoming, this paper presents a novel image dataset called CubiCasa5K, a large-scale floorplan image dataset containing 5000 samples annotated into over 80 floorplan object categories. The dataset annotations are performed in a dense and versatile manner by using polygons for separating the different objects. Diverging from the classical approaches based on strong heuristics and low-level pixel operations, we present a method relying on an improved multi-task convolutional neural network. By releasing the novel dataset and our implementations, this study significantly boosts the research on automatic floorplan image analysis as it provides a richer set of tools for investigating the problem in a more comprehensive manner. 

\end{abstract}

\noindent
\begin{tikzpicture}
  \footnotesize
  \draw [DarkGray,fill=Gray, line width=1pt,rounded corners=2mm]
    (0,0) rectangle
    (\columnwidth,1);
  \node at (.5\columnwidth,.5) {
    \parbox{.9\columnwidth}{\raggedright 
      \bf Data and code at: \href{https://github.com/CubiCasa/CubiCasa5k}{\color{black}{https://github.com/CubiCasa/CubiCasa5k}}
    }
  };
\end{tikzpicture}

\section{Introduction}
Floorplan image analysis or understanding has long been a research topic in automatic document analysis, a branch of computer vision. Floorplans are drawings to scale, they show the structure of a building or an apartment seen from above, and their purpose is to convey this structural information and related semantics to the viewer. Usual key elements in a floorplan are rooms, walls, doors, windows and fixed furniture, but they can cover more technical information as well, such as building materials, electrical wiring or plumbing lines. 

\begin{figure}[t]
  \centering
  \includegraphics[width=\linewidth]{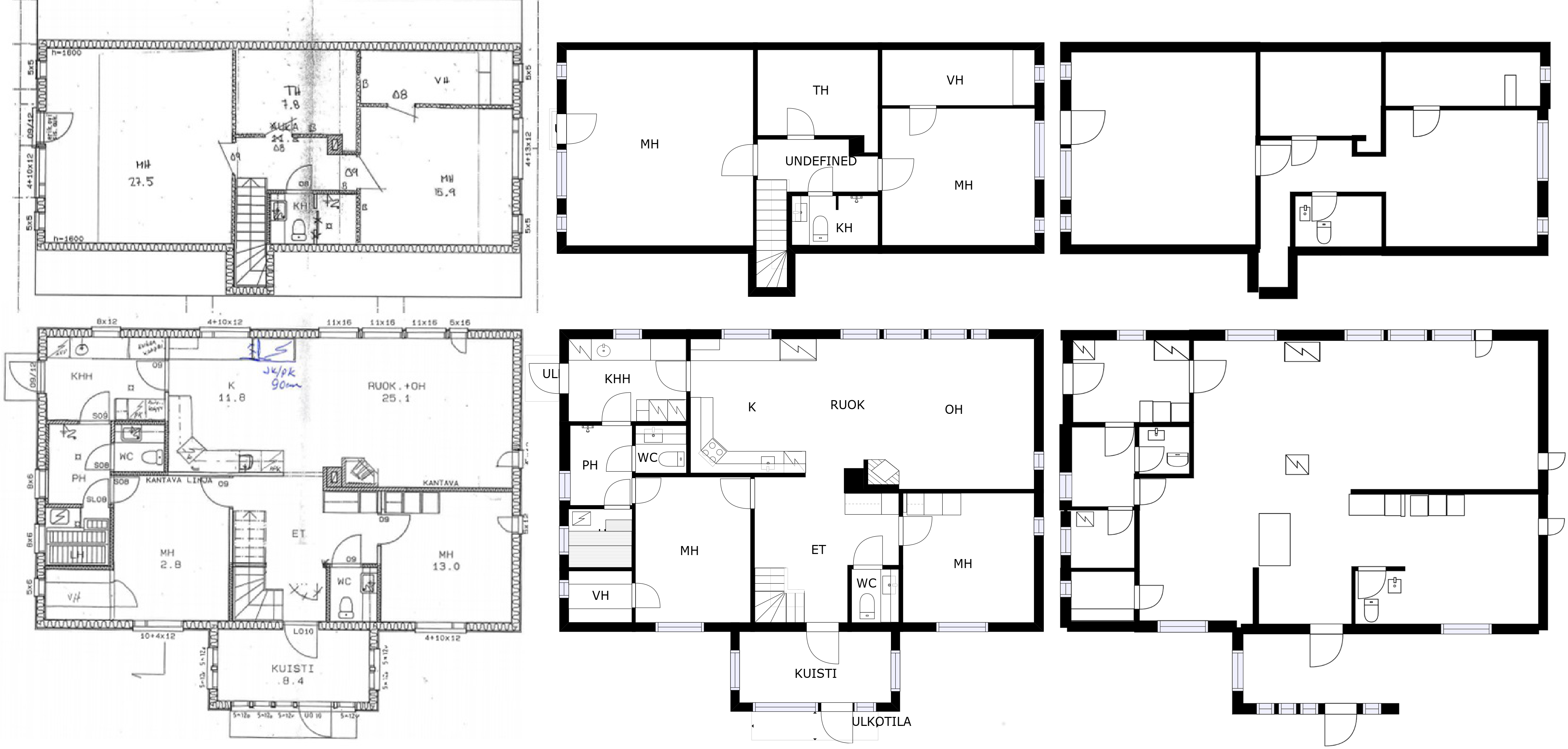}
  \caption{An example result using our dataset and pipeline. The images from left to right are original floorplan image, the SVG label, and the automatic prediction.}
  \label{fig:tranform}
\end{figure}

While floorplans are often initially drawn using a CAD software resulting in a vector-graphics format, for the usual use case that is in real estate economics, they are often rasterized before printing or publication in a digital media for marketing purposes, e.g. selling or renting. However, for the present day applications, such as 3D real estate virtual tours or floorplan-based 3D model creation,  this process is fatal as it discards all the structured geometric and semantic information, rendering effortless further utilization of these floorplan drawings troublesome.

Recovering the lost information from a rasterized floorplan image is not trivial. Current state of the art models in automatic floorplan image analysis are based on deep convolutional neural networks (CNNs). A promising training scheme is based on using only one network backbone together with several multi-task heads trained to recover the lost objects, structure and semantics. While the results are already promising \cite{Liu2017} the utilized datasets for training and benchmarking are still quite small (less than 1000~\cite{Liu2017}) compared to those datasets commonly applied in other mainstream computer vision domains, such as image classification (millions of images~\cite{2014ImageNet,Sun2017,2015JTF}) or image segmentation (tens of thousands images~\cite{Neuhold2017,zhou2017scene}). It is well known that deep learning models require large amounts of data to be effective and most likely increasing dataset sizes would always yield better results \cite{Neuhold2017,Sun2017}.

In this paper we propose a novel floorplan image dataset comprising out of 5000 images with dense and rich ground-truth annotations encoded all as polygons. The dataset covers three different floorplan image categories, namely high quality, high quality architecture and colorful. The annotations, generated by human experts, cover over 80 different floorplan element classes. The proposed dataset is over five times larger compared with the previously largest dataset \cite{Liu2017} and the annotation is more accurate as it includes precise shape and direction of objects. It also exhibits a larger variability in terms of the apartment type and in the style of the drawing. For a strong baseline, we present a fully automatic multi-task learning scheme inspired by the recent efforts reported in the literature. In detail, we use the recent `multi-task uncertainty loss' capable of deriving the weights for the losses of the network automatically. Our preliminary results indicate the method's great value in practice as it saves time from hyperparameter tuning in cases where the range of weights are totally unknown. We combine this loss with an encoder-decoder convolutional architecture and demonstrate state-of-the-art results in a previous floorplan analysis benchmark dataset. We release the proposed novel benchmarking dataset and our codes with trained models for easily reproducing the results of this paper.    

\section{Related Work}
As in many visual recognition problems, the research focus in 2D floorplan analysis has shifted from careful feature engineering to methods relying on learning from data. This shift is due to the ability to train larger and more powerful deep neural networks within a reasonable time~\cite{Lecun2015}. In our context, the breakthrough is~\cite{Liu2017}, which proposed an automatic floorplan parsing approach relying on CNNs. Instead of applying a bunch of low level image processing and consequent heuristics, a regular fully convolutional network was trained to label objects (rooms, icons, and openings) and to localise wall joints. The extracted low-level information was then fed in a post-processor to recover the original floorplan object items as 2D polygons. In~\cite{Liu2017}, the model was jointly optimized with respect to segmentation and localization tasks. The major finding was that deep neural networks can act as an effective precursor to the final post-processing heuristics to restore the floorplan elements, including their geometry and semantics. The method significantly improved the state of the art and has inspired recent research in the field. 

Parallel to~\cite{Liu2017}, a CNN-based method for parsing floorplan images using segmentation, object detection, and character recognition was proposed in~\cite{dodgeMVA2017}. The main difference to~\cite{Liu2017} is that the given tasks are all performed using isolated networks. The experiments performed in~\cite{dodgeMVA2017} on wall segmentation clearly demonstrated the superiority of a CNN-based approach compared with some traditional patch-based models utilizing standard shallow classifiers like support vector machines. In summary, the era of deep neural networks has given rise to significantly better methods for 2D floorplan analysis. According to~\cite{Liu2017,dodgeMVA2017}, especially fully convolutional CNNs have a huge potential in extracting accurate pixel-level geometric and semantic information that can be further utilized in later post-processing steps to construct more effective heuristics to restore the lost floorplan elements. 

The problem of constructing better CNNs for floorplan parsing boils down to two design choices that are related to the network architecture and the objective. The breakthrough in semantic segmentation research happened with the introduction of fully convolutional networks (FCNs)~\cite{Shelhamer2017}. A refined architecture for general purpose dense pixel-wise prediction is the U-net architecture with skip-connections~\cite{ronnebergerMICCAI2015}. As showed in~\cite{Liu2017}, the capacity can be further boosted by changing the plain convolutional layers in the top-down path to residual blocks~\cite{He2016}. This model, also known as the hourglass architecture~\cite{Bulat2016}, has proven effective in such dense problems as semantic segmentation~\cite{pohlen2016} and human pose estimation via heatmap regression~\cite{Bulat2016}. The final task is to choose the training objective. For a plain segmentation problem this is often a single cross-entropy loss or in heatmap regression a single euclidean loss layer. However, many problems in practice (like ours) can benefit from several objectives that are active during training, better known as multi-task learning~\cite{Caruana1997}. The success of using this approach is highly dependent on the additional hyperparameters which is the relative weighting between each task's loss. Kendall et al.~\cite{Kendall2017} proposed a simple solution to train this weighting in a multi-task setting composed of segmentation, depth estimation, and instance segmentation. In contrast to~\cite{Liu2017}, we apply the method of~\cite{Kendall2017} (revised in~\cite{Liebel}) to automatically tune the weighting between the tasks reducing the need for extensive hyperparameter tuning. Our results yield significant performance gains compared with the results reported in~\cite{Liu2017}.

To conclude, the current research on automatic floorplan conversion continues to lack representative large-scale datasets. At the moment, the largest annotated dataset publicly available contains less than 1K. The diversity of objects (e.g. different room and icon types) and consistency and accuracy in their annotation (e.g. thickness of walls) are both limited. This in turn implies that there are room for further studies to investigate the benefits of using larger datasets richer in their content for the training of deep CNNs. In this paper, we propose a dataset with 5K samples, to our knowledge the largest annotated floorplan dataset available.

\section{CubiCasa5K: A Novel Floorplan Dataset}
The CubiCasa5K dataset is a byproduct of an online, partially manual, floor plan vectorization pipeline~\footnote{\url{http://cubitool.cubi.casa.s3-website-us-west-2.amazonaws.com/?config=customize\&rl=2\&loc=na\&id=8000\&color=000000}}, mainly operating on real estate marketing material conversions from Finland region. Its main mission is to provide means for the research community to develop more accurate and automated models for real estate and other use cases.

The dataset consists of 5000 floorplans (with man-made annotations) that were collected and reviewed from a larger set of 15 000 mostly Finnish floorplan images. These are divided into three sub categories: high quality architectural, high quality and colorful, each containing 3732, 992 and 276 floorplans, respectively. To train powerful machine learning models, the dataset is randomly splitted into training, validation and test sets so that there are 4200, 400, and 400 floorplan in each of these, respectively. The annotations are in per-image SVG vector graphics format, and each of them contain the semantic and geometric annotations for all the elements appearing in the corresponding floorplan. 

\subsubsection{Annotations and their consistency.}
All samples of the proposed dataset have gone through an annotation pipeline resulting in a vectorized floorplan image with rich annotations. A sample input is always a raster scan (usually a scanned copy) generated from the original floor plan drawing. The annotation was done manually by human annotators who were educated to this task. Single image annotation took from 5 to 120 minutes, depending on the complexity and clearness of the source and amount of floors. 

Each floorplan has been annotated following an annotation protocol which describes the order for annotating the elements. That is to utilize all available information from the previously annotated elements (e.g. walls are boundaries to rooms) for a given floorplan. The annotation has been done using a special CAD tool tailored for drawing floorplans. To ensure annotation consistency, there is a QA process which has two stages. This process is applied to each annotated sample image. In detail, the applied QA process targets to control placement accuracy of the annotations as well as the correct label. The first round of the process is done by the annotator, who checks the annotated floor plan and reviews all the annotations, and finally corrects all possible errors. The second round is done by a different QA person who does the same checking procedure as the initial annotator and corrects errors slipped through the first round if any.

\subsubsection{Dataset statistics.} 
\begin{figure}[H]
\CenterFloatBoxes
\begin{floatrow}
\ffigbox
  {\includegraphics[width=0.8\linewidth]{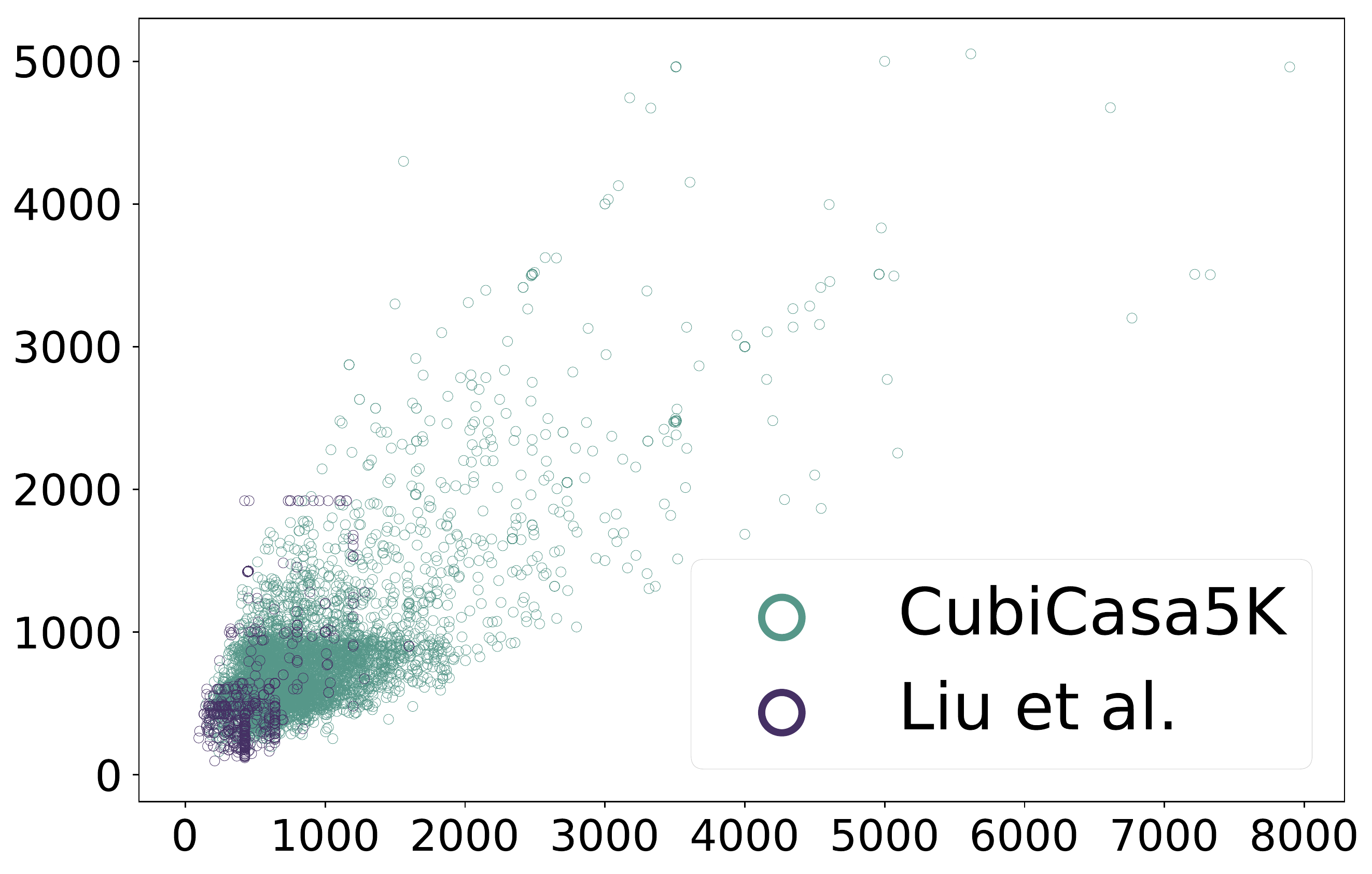}}
  {\caption{Source image resolutions of CubiCasa5K and Liu et al.~\cite{Liu2017}.\label{fig:scatter}}}
\killfloatstyle
\ttabbox
  {
  \scalebox{0.7}{
    \begin{tabular}{l | c | c | c | c }
                                & R-FP-500   & CVC-FP  & Liu et al.  & Cubi- \\
                                &  \cite{dodgeMVA2017}  &  \cite{Heras15a} & \cite{Liu2017}  & Casa5K \\
    \hline
    Images                      & 500       & 122           & 815           & 5000          \\
    Res                         & 56--1427  & 905--7383     & 96--1920      & 50--8000     \\ 
    Object                      & N/A       & 50            & 27            & 83            \\ 
    Room                        & N/A       & 1320          & 7466          & 68877      \\ 
    Icon                        & N/A       & 2345\footnote{Dataset included more icons labels, but without location or the polygon. We ignored these icons.}          & 15040         & 136676      \\ 
    Wall                        & N/A       & 6089          & 16139         & 147024      \\ 
    \end{tabular}}
  }
  {\caption{Metrics between available datasets.}\label{tab:dataset_comp}}
\end{floatrow}
\end{figure}

\begin{figure}[t]
   \centering
   \includegraphics[width=9cm]{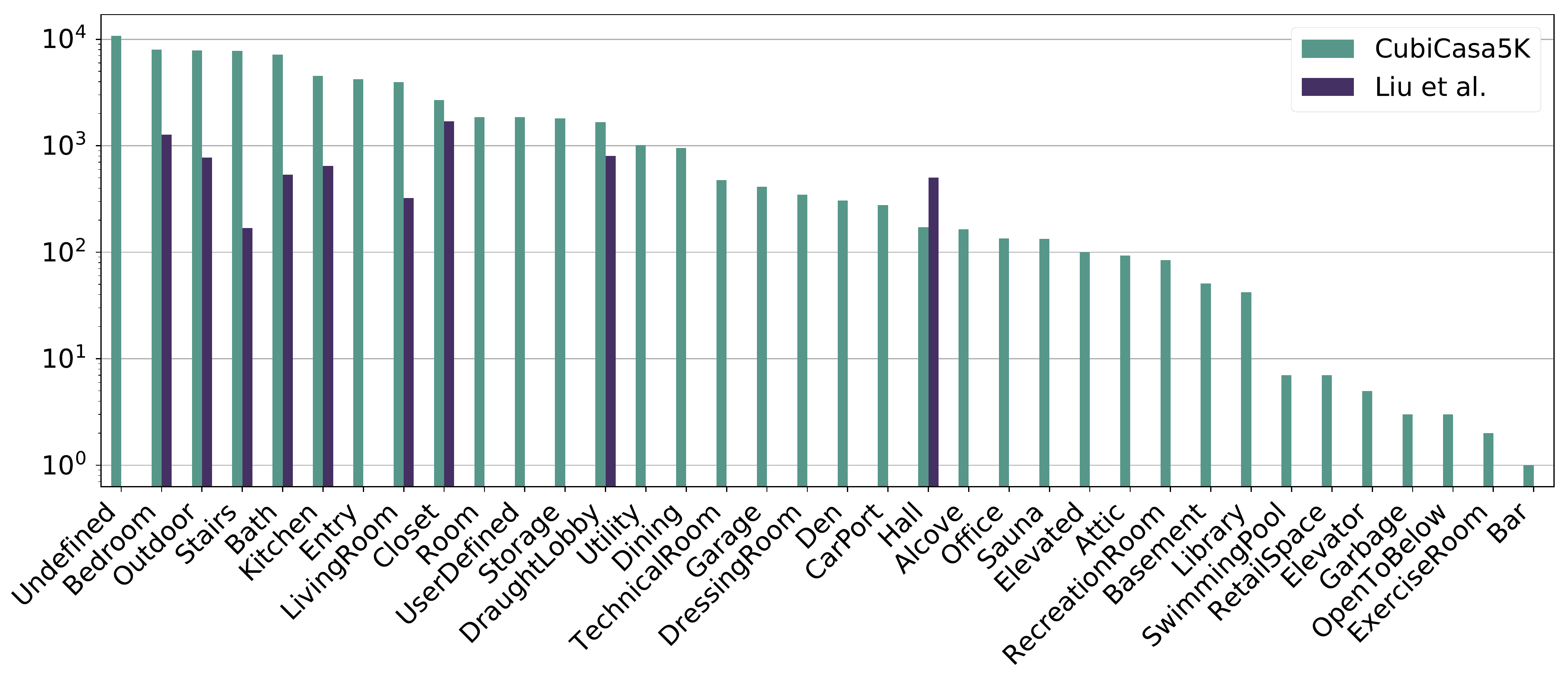}%
   \hspace{1in}
   \includegraphics[width=9cm]{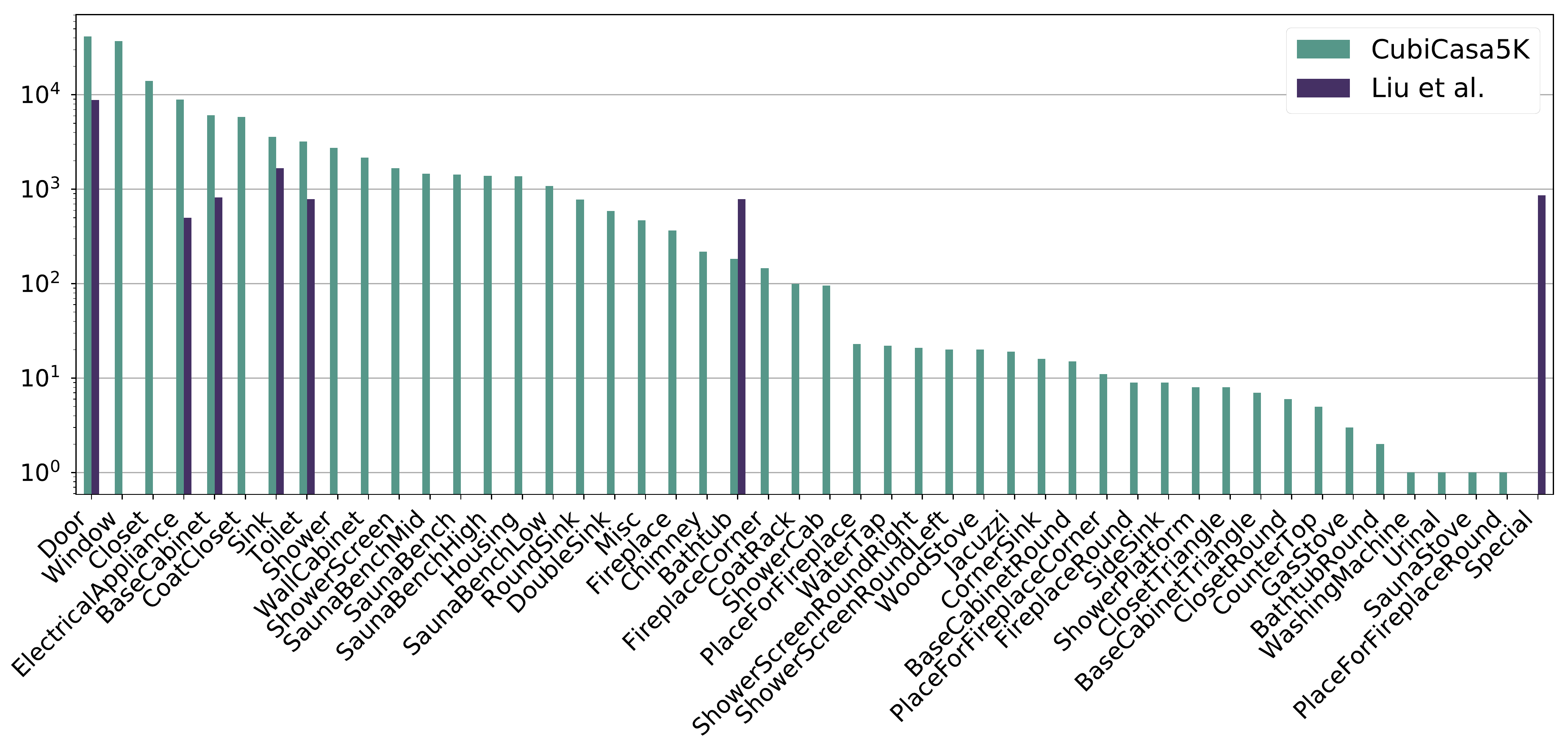}%
   \caption{Number of room (top) and icon (bottom) instances in our dataset. We made our best to match our labels with those in \cite{Liu2017}. For example, the `Entrance' room type in \cite{Liu2017} is considered as the `Draught Lobby' room type in CubiCasa5K, and the `PS' icon type in \cite{Liu2017} is considered as the `Electrical Appliance' in CubiCasa5K.} 
   \label{fig:room_count}
\end{figure}

\begin{figure}[htb]\scalebox{0.7}{
    \centering
    \includegraphics[width=.325\textwidth]{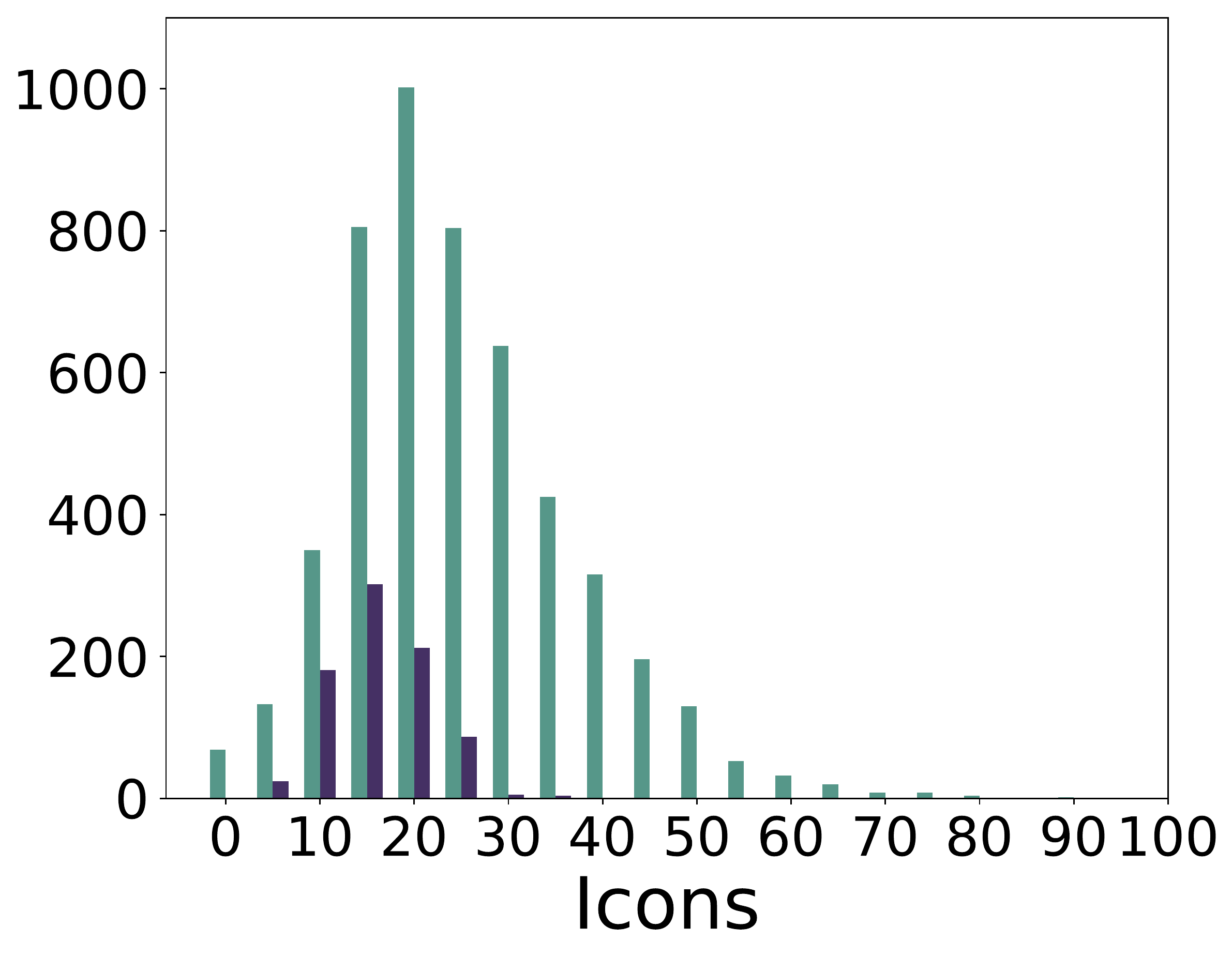}
    \includegraphics[width=.325\textwidth]{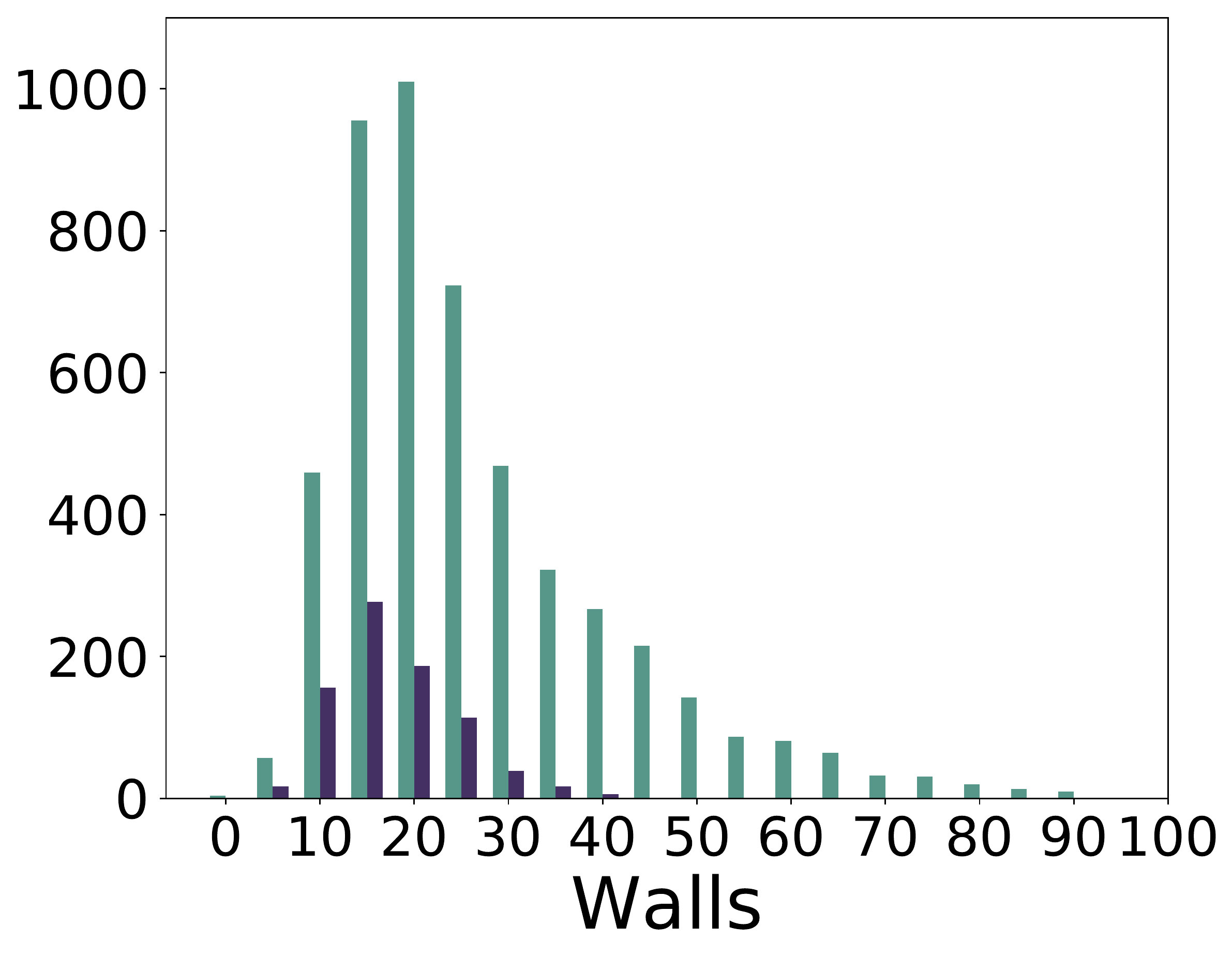}
    \includegraphics[width=.325\textwidth]{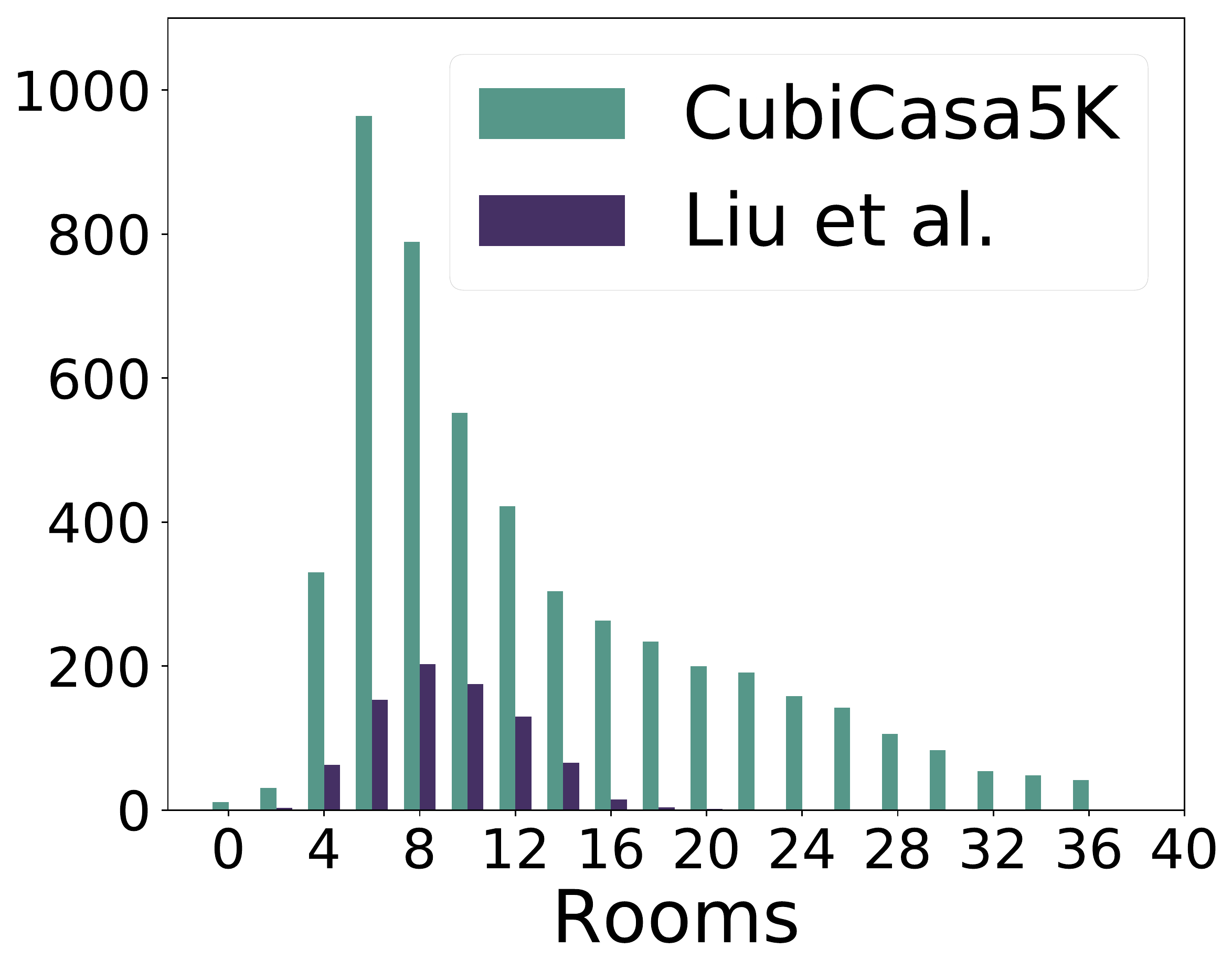}
    \caption{The frequency of floorplans containing certain number of annotated instances (from left to right) for icons, walls, and rooms.}
\label{fig:hist}}
\end{figure}

Fig. 2-4 provide statistical information about the CubiCasa5K dataset highlighting the aspects on the distribution of classes and complexity of the floorplan samples in relation to the dataset of~\cite{Liu2017}. Fig. \ref{fig:room_count} shows the distribution of ranked room and icon classes, respectively. In Fig. \ref{fig:hist}, we compare the frequency of images with a fixed number of annotated icons, walls, and rooms in the CubiCasa5K dataset and in the dataset of~\cite{Liu2017}. In Fig. \ref{fig:scatter} we report the distribution of the image resolution across the dataset. Finally, in Table \ref{tab:dataset_comp}, we further compare some key statistics to all existing annotated floorplan datasets. In the light of all this information, it can be conluded that the CubiCasa5K is currently the largest and the most versatile floorplan dataset publicly available.
 
\section{Our multi-task model}
Our task is to parse all common elements in an input 2D floorplan image. Following~\cite{Liu2017}, we rely on a network with outputs for two segmentation maps (one for different room types and one for different icon types) and a set of heatmaps to pinpoint wall junctions, icon corners, and opening endpoints (from now on, these three are referred to as interest points). Using the localized interest points, a set of heuristics is then applied to infer the geometry, i.e. location and dimensions, of all elements possibly appearing. Finally, the two segmentation maps are used to acquire the semantics, i.e. the types of rooms and icons. Our main contribution is in the latter step of the pipeline where we apply a trainable module~\cite{Kendall2017} for tuning the relative weighting between the multi-task loss terms.

\begin{figure}[t]
   \centering
   \includegraphics[height=3.1in]{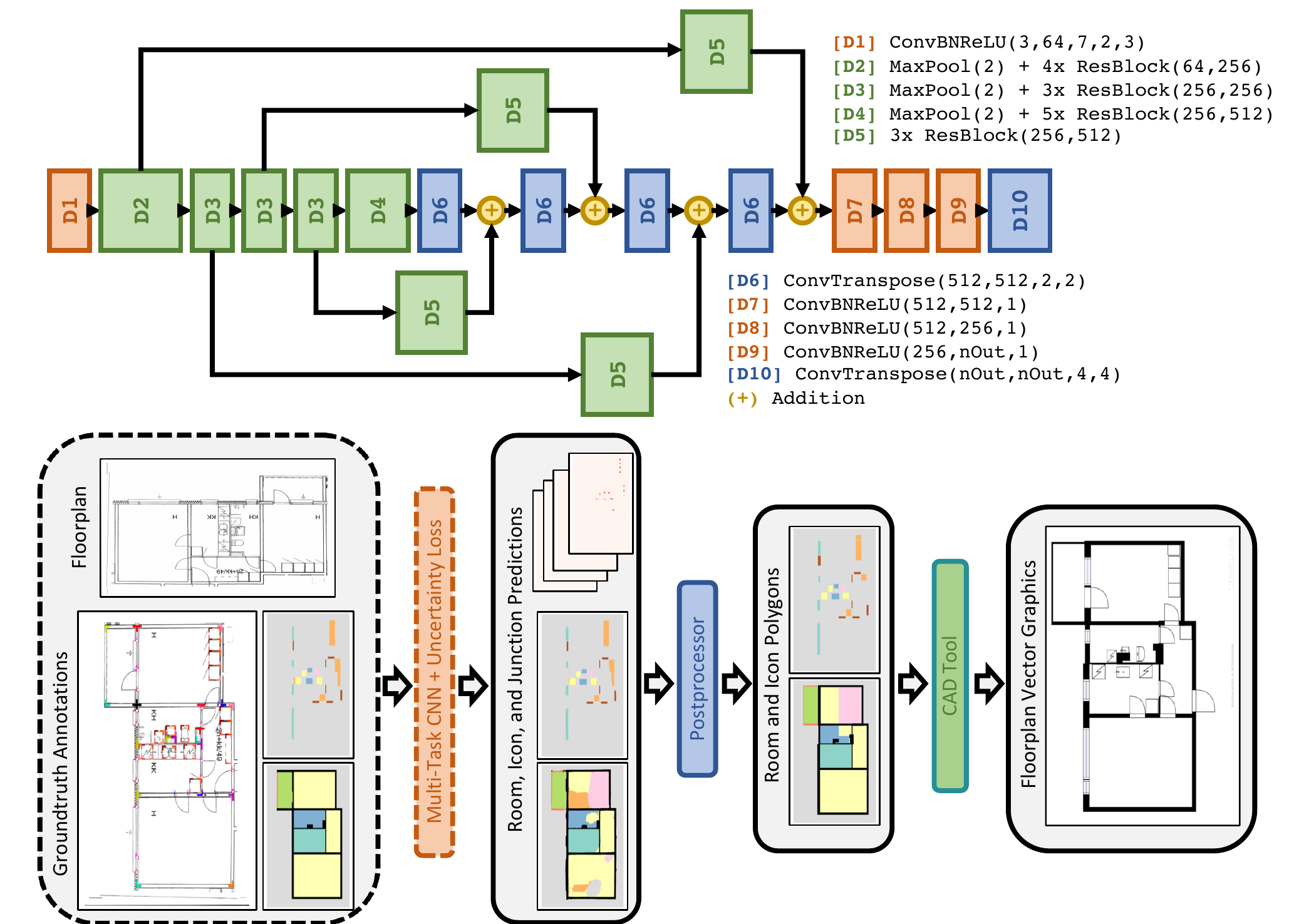}
   \caption{On top, the applied `hourglass network' architecture with blocks D1-10, where $x$ and $y$ of Resblock($x,y$) denote the number of channels in the very first and the number of outputs in the very latest block in the given `$n\times$ Resblock' sequence, respectively. ConvBNReLU($x, y, k, s, p$) follows the standard notation for a sequence of convolution, batch normalization and ReLU, where the arguments $k$, $s$, $p$ stand for the kernel size, stride and padding, respectively. ConvTranspose($x, y, k, s$) denotes transposed convolution with arguments $k$ and $s$ for the kernel size and stride, respectively. On bottom, the proposed pipeline for automatic floorplan parsing. The components outlined with a dashed line highlight our main contributions, i.e. a novel dataset and an improved multi-task model. } 
   \label{fig:hg_hourglass}
\end{figure}

\subsubsection{Network Architecture.}
We utilize the network architecture used in~\cite{Liu2017}, which is based on ResNet-152~\cite{He2016} pretrained with ImageNet~\cite{dengCVPR2009}. The organization of the layers is depicted on top in Fig.~\ref{fig:hg_hourglass} giving the details of each layer operation therein. Following~\cite{Liu2017}, the bulk of the network layers depicted in Fig.~\ref{fig:hg_hourglass} are initialized via training first on ImageNet~\cite{2014ImageNet} and then on the MPII Human Pose Dataset~\cite{Andriluka2014}. To have it tailored for the problem studied in this paper, some changes had to be made. Specifically, D1 was changed with respect to input channels (from 19 to 3), and the last two layers, namely D9 and D10, were both replaced to implement the required number of output channels for the two segmentation maps and 21 heatmaps. As a result, the three given layers (D1, D9, and D10) had to be randomly initialized. 

\subsubsection{Training Objective.}
In~\cite{Liu2017}, the approach relies on a shared representation for densely predicting semantic labels for pixels and regressing the locations for interest points. This means that there is a multi-task loss applied in the end of the network shown in Fig.~\ref{fig:hg_hourglass}. In detail, there are altogether 21 output maps for different interest points (wall junctions, icon corners, and opening endpoints). What is learned is a pixel-accurate location for all the interest points by means of separate heatmap regression tasks which all are based on a mean squared error (MSE) as the training objective. Besides this, the network outputs two segmentation maps. The first one is for segmenting background, rooms, and walls; and the second for segmenting the different icons and openings (windows and doors). The two segmentation tasks are both trained by applying a standard cross-entropy loss. In~\cite{Liu2017}, all these tasks were used to train the given shared representation in a multi-task fashion using a relative weighting fixed by hand.

A recent study by Kendall et al.~\cite{Kendall2017} shows that the relative weighting between the multi-task losses can be learnt automatically. This releases the developer from the difficult, time-consuming, and very expensive step of tuning the weights by hand. In specific, the weights are implicitly learnt via so called homoscedastic uncertainty terms that are predicted as an extra output for each task. The details can be found from~\cite{Kendall2017} and we proceed here directly to the final loss that is in our case given as $\mathcal{L}_{tot} = \mathcal{L_{H}} + \mathcal{L_{S}}$ where:

\begin{equation}\small
    \mathcal{L_{H}} = \sum_{i}\left [\frac{1}{2\sigma^2_i}\left \|\boldsymbol{y}_i - \boldsymbol{\mathrm{f}}_i^{\mathbf{W}}(\boldsymbol{x})  \right \|  + \log (1+\sigma_i)\right ],
\end{equation}
and
\begin{equation}\small
\mathcal{L}_{S} = -\sum_{k \in \left \{ rooms, icons  \right \}} \frac{1}{\sigma_{k}} \boldsymbol{y_{k}} \cdot \log \textup{softmax}(\boldsymbol{\mathrm{f}_{k}}^{\mathbf{W}}(\boldsymbol{x})) + \log \sigma_{k}.  
\end{equation}

$\mathcal{L_{H}}$ is for training the heatmap regressors and it is composed of a bunch of terms (as many as there are specific interest points to locate) minimized based a re-weighted MSE. The weighting is inversely proportional to a so called uncertainty parameter $\sigma_{i}$ that is learnt during training. The terms $\log(1+\sigma_i)$~\cite{Liebel} act as a regularizer to avoid the trivial solution. Furthermore, by summing one before taking logs, we enforce it to be always positive~\cite{Liebel}. $\mathcal{L_{S}}$, in turn, is for the segmentation part and it is composed of two cross-entropy terms, in this case for room and icon segmentation tasks, to be minimized. The weighting in this case appears without squaring~\cite{Kendall2017}. Based on our experimental findings, the regularizer $\log\sigma_{k}$ stays positive whole the time during training. 

\subsubsection{Post-processing.}
To generate the final vector graphics equivalent representation of the input rasterized floorplan, the outputs of the multi-task CNN are dispatched to a post-processor which consists of four steps. The target is to extract all floorplan elements (walls, rooms, icons, and openings) present in the given input in a format precisely encoding their location, dimensions, and the category label.

The post-processor starts with inferring wall polygons. In detail, the procedure starts with the same step as in~\cite{Liu2017}: the junctions are pair-wisely connected based on their joints' orientation, i.e. if there are two junctions vertically/horizontally aligned (with a possible few pixel misalignment) in a manner where both have a joint facing each other. The procedure results in a wall skeleton which is next pruned based on the wall segmentation. Finally, the width of the wall is inferred by sampling along the wall lines and inspecting the intensity profile of the wall segmentation map. 

The location and dimensions of rooms are partly inferred based on the wall junctions. In detail, we search for all junction triplets that span a rectangular area that does not contain any junctions. This results in a cell gridding of the floorplan interiors. The resulting cells are then labeled according to a voting mechanism based on the room segmentation map. Finally, all neighboring cells are merged if an only if there is no fully separating walls between them and they share the same room label. The procedure for restoring icons is very similar to rooms extraction, but instead of the wall junction heatmaps, we utilize the triplets from the maps responsible for the icon corner heatmap prediction. 

Finally, doors and windows are inferred. This is done by connecting the two vertically/horizontally aligning opening endpoints using the predictions from the corresponding heatmaps. The label is again derived based on the segmentation maps. The width of the opening is the same as the wall polygon. All such opening endpoints that do not fall inside the wall segmentation are rejected. 

\section{Results}
In this section we introduce the evaluation metrics and the obtained results. Prior to presenting the baseline results for our novel CubiCasa5K dataset, we validate our method on the same dataset used in~\cite{Liu2017}.

\subsubsection{Preliminary Experiments.}
Following \cite{Liu2017}, the network was initialized with the weights of the human pose network of \cite{Bulat2016} (trained on ImageNet and MPII). Those layers that had to be replaced (see Sect. 4) were initialized randomly. We trained the network with uncertainty driven task weighting for 400 epochs with a batch size of 20. The data augmentations were 90 degree rotations, color jitter, and randomly selecting between crop and scaling to 256x256 with zero padding. We used the Adam optimizer with an initial learning rate of \num{1e-3}, $\epsilon=\num{1e-8}$ and $\beta$ values of 0.9 and 0.999. We used a scheduler that reduced the learning rate with a factor of 0.1 if no improvement were noticed during the previous 20 epochs based on the validation loss. After dropping, the training was then continued from the phase that had yielded the best validation loss up to that point. Finally, the best model was selected based on the validation loss. Based on our experiments, the learning rate had to be dropped only once and the training seemed to converge just before the epoch number 300. The training took three hours on a Nvidia GeForce GTX TitanX GPU card.

For evaluating our model we used the same evaluation setup as in~\cite{Liu2017}. As can be seen in Table \ref{tab:furukawa_results} we significantly improve the results presented in \cite{Liu2017}. We further applied a test-time augmentation (TTA) scheme in which the final predictions were based on feeding the same image four times to the same network, each time rotating it 90 degrees. The final prediction was based on the mean of the four predictions. As it can be seen, the augmentation seems to be beneficial in both cases, with and without integer programming (IP). 
\begin{table}[H]
\caption{Evaluation results on the dataset proposed in~\cite{Liu2017}.}
\centering
\scalebox{0.7}{
\begin{tabularx}{\textwidth}{ l | X | X | X | X | X | X | X | X |}
    
    \multirow{2}{*}{Method}                      & \multicolumn{2}{c|}{Junction}& \multicolumn{2}{c|}{Opening} & \multicolumn{2}{c|}{Icon}    & \multicolumn{2}{c|}{Room}    \\
                                & acc          & recall           & acc           & recall           & acc           & recall           & acc           & recall           \\ \hline
    \cite{Liu2017}                   & 70.7          & \textbf{95.1} & 67.9          & 91.4          & 22.3          & 77.4          & 80.9          & 78.5          \\ \hline
    \cite{Liu2017} + IP                 & 94.7          & 91.7          & 91.9          & 90.2          & 84.0          & 74.6          & 84.5          & 88.4          \\ \hline
    \cite{Liu2017} (our eval)     & 75.5          & 90.0          & 74.6          & 91.8          & 25.3          & 79.9          & 84.6          & 83.5          \\ \hline
    \cite{Liu2017} + IP (our eval)   & 92.9          & 86.6          & 92.3          & 90.6          & 86.8          & 78.5          & 89.9          & 88.3          \\ \hline
    best reproduced from \cite{Liu2017} & 75.6 & 88.4 & 72.5 & 89.3 & 23.1 & 73.2 & 85.9 & 83.3 \\ \hline
    best reproduced from \cite{Liu2017} + IP & 93.1 & 84.5 & 91.4 & 88.1 & 80.7 & 72.1 & 89.1 & 87.1 \\  \hline
    Ours                        & 82.4          & 92.0          & 82.3          & 93.3          & 34.6          & \textbf{88.3} & 90.0          & 87.6          \\ \hline
    Ours (TTA)          & 90.2          & 91.9          & 89.6          & \textbf{93.9} & 46.1          & 88.0          & 91.5          & 88.0          \\ \hline
    Ours + IP                     & 94.1          & 89.6          & 93.2          & 92.6          & 92.9          & 87.7          & 91.7          & \textbf{90.8} \\ \hline
    Ours (TTA) + IP   & \textbf{95.0} & 89.7          & \textbf{94.5} & 92.9          & \textbf{93.6} & 87.3          & \textbf{92.2} & 90.2          \\ \hline
\end{tabularx}}
\label{tab:furukawa_results}
\end{table}
We noticed an error in the original annotations\footnote{Please see the target labels corrected by us on our project webpage.} of the dataset proposed in~\cite{Liu2017}. After re-evaluating the model of~\cite{Liu2017} using our fixes, we noticed that the performance of~\cite{Liu2017} (see `our eval') is actually better than originally reported in~\cite{Liu2017}. We further trained the model (`best reproduced from~\cite{Liu2017}') by following the details reported in the original paper, and the results were more or less similar. Finally, we compared our best model (`Ours') without test-time augmentations to the `our eval' version of~\cite{Liu2017}, and as it can be seen, our model is clearly better.

\subsubsection{CubiCasa5K Experiments.}
In the present experiment utilizing the CubiCasa5K dataset, some original room types and icon types are coupled so that our targets cover altogether 12 room and 11 icon classes (see the chosen classes in Table \ref{tab:cubicasa_class_results400} and further details from the project website). As for other details, the network contains the same heatmap regression layers and is trained using the same objective as in the previous experiment. However, the following adjustments to the training scheme were done: We started the training with the pretrained weights as a result of the previous experiment using ImageNet, MPII, and Liu et al.~\cite{Liu2017} datasets. We trained the first 100 epochs with the same augmentations given in Sect. 5.1. After that, we continued training with the best weights up to that point (according to the losses on the validation set) by first initializing optimizer parameters to their starting values and then dropping the augmentation that resizes the image to 256x256. We then trained the network for 400 epochs which resulted in convergence.

Following the common practice in the art of semantic segmentation~\cite{Shelhamer2017,zhou2017scene}, we report the results using three evaluating metrics, namely the \emph{overall accuracy} indicating the proportion of correctly classified pixels, and the \emph{mean accuracy} for the proportion of correctly classified pixels averaged over all the classes. Finally, we report the \emph{mean intersection over union} (IoU) which indicates the area of the overlap between the predicted and ground truth pixels, averaged over all the classes. We further report the results with respect to the raw segmentations and polygonized (P) instances, i.e. after the post-processing step. We take a different approach to~\cite{Liu2017} for model evaluation as we believe the problem of floorplan parsing is very close to the problem of semantic segmentation.

We report the performance with respect to described metrics in Table \ref{tab:cubicasa_results}. According to the results, the raw segmentation test scores are clearly better than the ones based on polygonized segmentation instances. The main reason is that if wall or icon junctions are missed or are not correctly located the polygons can not be created regardless the quality of the segmentation. In Table \ref{tab:cubicasa_class_results400}, we further report the class-specific IoUs and accuracies with respect to all room and icon classes used in this study. Fig.1 illustrates an example result from our pipeline.

\section{Conclusions}
In this paper, we have proposed a novel floorplan image dataset called CubiCasa5K. Compared to other existing annotated floorplan datasets, our dataset is over 5$\times$ larger and more varied in its annotations which cover over 80 floorplan object categories. Together with the novel dataset, we provided baseline results using an improved multi-task convolutional neural network yielding state-of-the-art performance.

For future directions, we plan to integrate the object detector used in~\cite{dodgeMVA2017} as one of the tasks into our pipeline. It would be also of interest to try the method of~\cite{Acuna_2018_CVPR} to directly infer floorplan elements as polygons.

\begin{table}[t]
\caption{Evaluation results of the CubiCasa5K dataset.} \label{tab:cubicasa_results}
\scalebox{0.7}{
    \centering
      \begin{tabular}{ l | c | c | c | c | c | c | }
        \multirow{ 2}{*}{} & \multicolumn{2}{c|}{Overall Acc} & \multicolumn{2}{c|}{Mean Acc} & \multicolumn{2}{c|}{Mean IoU} \\
                                & val & test & val & test & val & test \\ \hline
        Rooms                   & 84.5 & 82.7 & 72.3 & 69.8 & 61.0 & 57.5 \\ \hline
        Room\textsubscript{P}   & 79.0 & 77.3 & 64.2 & 61.6 & 52.4 & 49.3 \\ \hline
        Icons                   & 97.8 & 97.6 & 62.8 & 61.5 & 56.5 & 55.7 \\ \hline
        Icons\textsubscript{P}  & 97.0 & 96.7 & 94.8 & 45.3 & 43.7 & 41.6 \\ \hline
      \end{tabular}
}
\end{table}

\newcommand*\rot{\rotatebox{90}}
\newcommand*\OK{\ding{51}}
\begin{table}[t]

\caption{Class-specific results on the CubiCasa5K dataset.}
\label{tab:cubicasa_class_results400}
\scalebox{0.7}{
\begin{tabular}{ll|l|l|l|l|l|l|l|l|l|l|l|l||l|l|l|l|l|l|l|l|l|l|l|}
                         &      & \multicolumn{12}{c||}{Rooms}       & \multicolumn{11}{c|}{Icons} \\ 
\multicolumn{2}{l|}{}&\rot{Background}&\rot{Outdoor}&\rot{Wall}&\rot{Kitchen}&\rot{Living Room}&\rot{Bedroom}&\rot{Bath}& \rot{Hallway}&\rot{Railing}&\rot{Storage}&\rot{Garage}& \rot{Other rooms} & \rot{Empty}& \rot{Window} & \rot{Door} & \rot{Closet} & \rot{Electr. Appl.} & \rot{Toilet} & \rot{Sink} & \rot{Sauna bench}& \rot{Fire Place} & \rot{Bathtub} & \rot{Chimney} \\ \hline
\multirow{2}{*}{IoU}                    & val  & 88.3 & 62.6 & 74.0 & 71.6 & 68.2 & 80.0 & 66.5 & 63.0 & 29.0 & 51.6 & 32.2 & 45.2 & 97.8 & 67.3 & 56.7 & 69.8 & 67.4 & 65.3 & 55.5 & 72.7 & 38.4 & 16.7 & 13.8 \\ \cline{2-25} 
                                        & test & 87.3 & 64.4 & 73.0 & 65.0 & 66.6 & 74.2 & 60.6 & 55.6 & 23.6 & 44.8 & 33.7 & 41.4 & 97.6 & 66.8 & 53.6 & 69.2 & 66.0 & 62.8 & 55.7 & 67.3 & 36.2 & 26.7 & 11.2 \\ \hline
\multirow{2}{*}{IoU\textsubscript{P}}   & val  & 80.3 & 44.9 & 50.8 & 66.3 & 63.5 & 74.6 & 62.3 & 60.4 & 7.8  & 46.6 & 28.7 & 42.4 & 96.9 & 47.7 & 46.8 & 62.5 & 60.6 & 60.8 & 43.4 & 52.1 &  4.7 &  0.0 &  5.7 \\ \cline{2-25} 
                                        & test & 79.2 & 48.5 & 47.9 & 58.3 & 62.2 & 68.7 & 56.6 & 53.5 & 5.8  & 40.8 & 30.0 & 39.8 & 96.7 & 40.9 & 41.2 & 63.3 & 59.8 & 56.2 & 45.5 & 48.0 &  3.1 &  0.0 &  2.9 \\ \hline
                            \hline
\multirow{2}{*}{Acc}                    & val  & 95.3 & 71.4 & 86.5 & 86.3 & 82.2 & 91.7 & 81.7 & 75.8 & 34.6 & 60.7 & 42.1 & 59.5 & 99.3 & 75.2 & 63.3 & 77.7 & 76.6 & 72.5 & 67.4 & 80.4 & 45.2 & 19.1 & 14.4 \\ \cline{2-25} 
                                        & test & 93.6 & 77.7 & 85.8 & 79.9 & 82.6 & 86.2 & 73.4 & 71.2 & 28.7 & 53.9 & 47.2 & 57.1 & 99.3 & 73.7 & 59.8 & 77.6 & 75.7 & 68.4 & 66.1 & 74.2 & 40.4 & 30.1 & 11.7 \\ \hline
\multirow{2}{*}{Acc\textsubscript{P}}   & val  & 93.5 & 56.9 & 55.8 & 82.9 & 80.4 & 89.4 & 79.1 & 72.8 & 8.2  & 57.0 & 37.1 & 57.3 & 99.4 & 52.3 & 53.5 & 67.5 & 68.3 & 67.5 & 52.0 & 54.8 &  4.8 &  0.0 &  5.7 \\ \cline{2-25} 
                                        & test & 92.9 & 60.6 & 52.8 & 74.2 & 81.3 & 83.6 & 69.9 & 68.3 & 6.1  & 49.4 & 43.8 & 55.8 & 99.4 & 44.8 & 47.4 & 69.1 & 67.6 & 60.8 & 52.9 & 49.7 &  3.1 &  0.0 &  2.9 \\ \hline
\end{tabular}}
\end{table}

%
%
%
\bibliographystyle{splncs04}
\bibliography{sources}
%




\end{document}